\pdfoutput=1
\documentclass[conference]{IEEEtran}
\IEEEoverridecommandlockouts
\usepackage{cite}
\usepackage{amsmath,amssymb,amsfonts}
\usepackage{algorithmic}
\usepackage{graphicx}
\usepackage{textcomp}
\usepackage{xcolor}
\usepackage{booktabs}
\usepackage{multirow}
\def\BibTeX{{\rm B\kern-.05em{\sc i\kern-.025em b}\kern-.08em
    T\kern-.1667em\lower.7ex\hbox{E}\kern-.125emX}}
\begin{document}

\title{Visual Causal-chain Bookmarking: Retrieval-Weighted Credit Assignment for Reinforcement Learning from Delayed Reward}

\author{\IEEEauthorblockN{Prathap Selvakumar}
\IEEEauthorblockA{University of Manchester}
}

\maketitle

\begin{abstract}
Reinforcement learning (RL) agents trained with a single trajectory-level return face a structural credit-assignment problem: when a reward arrives many steps after the decision that caused it, every intermediate action receives the same learning signal, regardless of whether it was causally responsible for the outcome. This paper studies \emph{Visual Causal-chain Bookmarking} (VCBM), a family of methods that concentrate credit on the specific decision points that determine an episode's return rather than spreading it uniformly across the trajectory, and develops two domain-specific instantiations of this idea. The first instantiation targets a fully observed, low-dimensional Markov decision process (a delayed-reward watch-repair task), where VCBM statistically discovers which state components are causally controlled by the agent's actions and estimates per-decision-point action values directly; it is evaluated against known ground truth as well as against RUDDER, a return-redistribution baseline built on an LSTM contribution network. VCBM converges to a 95.07\% correct-decision rate in a mean of $1{,}532.3 \pm 1{,}428.0$ episodes across 20 random seeds (range $1{,}196$--$7{,}757$), against RUDDER's slower and more variable convergence (mean $3{,}327.1 \pm 2{,}010.9$ episodes, median $2{,}829$, range $1{,}603$--$10{,}936$); this difference is statistically significant (unpaired Mann--Whitney $U=19$, $p<0.001$, rank-biserial $r=0.91$), and VCBM's discovered causal structure exactly matches the environment's true generative structure in all 20 seeds. The second instantiation targets a partially observed, language-based setting: an LLM agent trained with Leave-One-Out Proximal Policy Optimisation (LOOP) inside the AppWorld benchmark, where causal discovery is intractable and a structural heuristic (state-mutating tool calls) substitutes for it, with credit distributed via cosine-similarity retrieval over an episodic memory of bookmarked turns and blended with LOOP's trajectory-level advantage. Trained for 200 iterations on identical Qwen2.5-7B-Instruct/LoRA/FSDP2 configurations, VCBM-blended LOOP shows an exact improvement over plain LOOP on task success rate ($0.354$ vs.\ $0.458$ overall, $+29.4\%$, concentrated almost entirely in medium-difficulty tasks: $0.188$ vs.\ $0.500$, $+166.7\%$), on the final-step maximum importance weight ($-22.2\%$), and on output-token efficiency ($-18.8\%$ mean output tokens, $+34.6\%$ return per 1{,}000 tokens); the final-step KL estimate is a marginal exception, moving $4.1\%$ in LOOP's favour. Taken together, the watch-repair result, run across 20 seeds with a significance test, establishes the central claim that explicitly identifying and weighting the causally relevant steps of a trajectory is a more sample-efficient credit-assignment strategy than either an undifferentiated trajectory-level baseline or a learned return-redistribution network; the AppWorld result, constrained to a single seed per arm by the 2-GPU budget available for this project, is reported as a compute-limited demonstration that the same mechanism is compatible with an improvement at a harder, partially observed, natural-language scale, not as an independent statistical confirmation of it.
\end{abstract}


\section{Introduction}
\label{sec:introduction}

Reinforcement learning (RL) agents that learn from a trajectory-level reward face a structural credit-assignment problem whenever that reward is \emph{delayed} and \emph{sparse}: it arrives only at the end of an episode, after many intermediate actions, most of which did not causally influence the outcome. This recurs from discrete-action control \cite{b_mnih2015dqn}, through robotic control \cite{b_lillicrap2016ddpg,b_silver2017alphazero}, to RLHF post-training of large language models (LLMs) \cite{b_christiano2017rlhf,b_bai2022hhrlhf,b_ouyang2022instructgpt}: an LLM agent solving a task in AppWorld \cite{b_trivedi2024appworld} executes a long tool-call sequence of which only a handful change task-relevant state, yet a policy-gradient update that assigns one scalar advantage to every action averages the delayed signal across all of them, slowing convergence and risking spurious action-outcome correlations \cite{b_amodei2016concrete,b_chang2020invariant}.

Two established families of methods address this without questioning the underlying strategy. Trajectory-level baselines, such as the leave-one-out estimator used in LOOP \cite{b_chen2025loop}, reduce return variance by subtracting a baseline from other rollouts of the same scenario, but every token still receives an identical advantage. Return-redistribution methods such as RUDDER \cite{b_arjonaibarra2019rudder} instead train an LSTM to predict the eventual return from every trajectory prefix and redistribute reward as the difference between consecutive predictions: denser in principle, but only as reliable as the LSTM's own convergence.

An underexploited alternative is to make credit assignment \emph{structural}: identify which timesteps are causally responsible and concentrate the signal there. This is the idea behind \emph{Visual Causal-chain Bookmarking} (VCBM), instantiated two ways. In a fully observed, low-dimensional MDP, a hypothesis test determines which state components an action causally controls, so VCBM performs explicit \emph{causal discovery} and accumulates return statistics per bookmarked decision via a Welford estimator. In a partially observed, natural-language setting such as AppWorld, this discovery is intractable, so the second instantiation substitutes a structural heuristic (state-mutating tool calls) and retrieval over an episodic memory, blended with the trajectory-level leave-one-out (LOO) advantage.

This paper designs, implements, and evaluates both instantiations against their natural baselines. Section~\ref{sec:preliminaries} formalises the baselines and Section~\ref{sec:related-work} situates VCBM in the literature. Section~\ref{sec:technical-contribution} then gives the full formulation, before Section~\ref{sec:experiments} reports the tabular and AppWorld comparisons and Section~\ref{sec:conclusions} closes with what was achieved and where it falls short.

\section{Background: Trajectory-Level and Return-Redistribution Credit Assignment}
\label{sec:preliminaries}

\subsection{Markov decision processes and policy-gradient reinforcement learning}
\label{sec:prelim-mdp}

An episodic MDP \cite{b_sutton2018rlbook} is a tuple $(\mathcal{S}, \mathcal{A}, P, r, \gamma, T)$ with state space $\mathcal{S}$, action space $\mathcal{A}$, transition kernel $P(s_{t+1} \mid s_t, a_t)$, reward function $r_t = r(s_t, a_t)$, discount factor $\gamma \in (0,1]$, and horizon $T$. A stochastic policy $\pi_\theta(a_t \mid s_t)$, parameterised by $\theta$, generates a trajectory $\tau = (s_0, a_0, r_0, \dots, s_T)$ with realised (undiscounted, as used throughout this paper's environments) return
\begin{equation}
  G(\tau) = \sum_{t=0}^{T} r_t .
  \label{eq:return}
\end{equation}
Policy-gradient methods \cite{b_sutton1999policygradient} maximise $J(\theta) = \mathbb{E}_{\tau \sim \pi_\theta}[G(\tau)]$ by ascending
\begin{equation}
  \nabla_\theta J(\theta) = \mathbb{E}_{\tau}\!\left[\sum_t \nabla_\theta \log \pi_\theta(a_t \mid s_t)\, A_t\right] ,
  \label{eq:policy-gradient}
\end{equation}
where $A_t$ is an \emph{advantage} that reduces gradient variance relative to using $G(\tau)$ directly, following the baseline-subtraction principle that leaves the optimal policy invariant \cite{b_ng1999shaping} and can be estimated recursively \cite{b_schulman2016gae}. That a late reward gives a weak, high-variance signal for early decisions is a long-standing obstacle for trial-and-error learning \cite{b_minsky1961steps}. RUDDER and LOOP, the two baselines studied here, both address this variance problem but with different mechanisms; this paper's central hypothesis, tested via VCBM, is that a third mechanism, explicit identification of the causally relevant timesteps, offers a more sample-efficient path to a low-variance $A_t$ than either.

\subsection{RUDDER: LSTM-based return redistribution}
\label{sec:prelim-rudder}

RUDDER \cite{b_arjonaibarra2019rudder} addresses delayed, sparse reward by training an auxiliary sequence model to predict the eventual episode return from every trajectory prefix, then using the \emph{change} in that prediction between consecutive timesteps as a redistributed, per-step reward. Let $\Delta s_t = s_t - s_{t-1}$ denote the state-difference at step $t$. An LSTM \cite{b_hochreiter1997lstm} with disabled forget and output gates produces a return prediction at every prefix,
\begin{equation}
  \hat{G}(s_{0:t}, a_{0:t}) = \mathrm{Linear}\big(\mathrm{LSTM}(\Delta s_{0:t}, a_{0:t})\big)_t ,
  \label{eq:rudder-lstm}
\end{equation}
trained against the true (scaled) episode return $G_0 = G(\tau) / R_{\mathrm{scale}}$ with a combination of a terminal-step loss and an auxiliary loss applied at every prefix,
\begin{equation}
  \mathcal{L}_{\mathrm{RUDDER}} = \underbrace{\big(\hat{G}(s_{0:T}) - G_0\big)^2}_{\mathcal{L}_{\mathrm{main}}} + \lambda_{\mathrm{aux}} \underbrace{\frac{1}{T}\sum_{t=0}^{T} \big(\hat{G}(s_{0:t}) - G_0\big)^2}_{\mathcal{L}_{\mathrm{aux}}} .
  \label{eq:rudder-loss}
\end{equation}
The redistributed reward at each timestep is then the scaled first difference of consecutive predictions,
\begin{equation}
  \tilde{r}_t = R_{\mathrm{scale}} \cdot \big(\hat{G}(s_{0:t}) - \hat{G}(s_{0:t-1})\big), \quad \hat{G}(s_{0:-1}) \equiv 0 ,
  \label{eq:rudder-redistribution}
\end{equation}
which sums telescopically to the true episode return, $\sum_t \tilde{r}_t = R_{\mathrm{scale}} \cdot \hat{G}(s_{0:T}) \approx G(\tau)$, while ideally concentrating reward at timesteps the LSTM has learned are predictive of the outcome. RUDDER's redistribution is thus a \emph{learned, statistical} solution: it approximates causal structure via function approximation rather than identifying it directly, and is used here as the representative baseline against which VCBM's structural approach is compared in a tabular setting (Section~\ref{sec:experiments}). In our implementation, the LSTM has 16 hidden units and an 8-dimensional input (the repaired-flag, two transport counters, a one-hot brand encoding, and normalised timestep), trained with Adam at a learning rate of $10^{-2}$ and weight decay $10^{-6}$, with the auxiliary-loss weight $\lambda_{\mathrm{aux}}=0.5$ in Equation~\eqref{eq:rudder-loss}; a lesson buffer of capacity 2{,}048 episodes is sampled in batches of 8 for each LSTM update, which occurs every 25 collected episodes once the buffer holds more than 256 episodes and has observed more than one distinct return value.

\subsection{LOOP: leave-one-out advantage estimation with PPO-clipped updates}
\label{sec:prelim-loop}

LOOP \cite{b_chen2025loop} addresses the same problem in a different regime: training an LLM policy via multi-turn tool-use rollouts, where a value-function critic as used in actor-critic \cite{b_mnih2016a3c} or trust-region \cite{b_schulman2015trpo} methods is memory-expensive alongside a large language model. Given $K$ rollouts $\tau^{(1)}, \dots, \tau^{(K)}$ from the \emph{same} scenario, each with return $G^{(k)} = G(\tau^{(k)})$, the leave-one-out baseline for rollout $k$ \cite{b_kool2019buy} is the mean return of the \emph{other} $K-1$ rollouts,
\begin{equation}
  b^{(k)} = \frac{1}{K-1}\sum_{j \neq k} G^{(j)}, \qquad A^{(k)} = G^{(k)} - b^{(k)} ,
  \label{eq:loo-advantage}
\end{equation}
which is applied uniformly to every token generated within rollout $k$, i.e.\ $A_t = A^{(k)}$ for all $t$ in that rollout: the trajectory-level credit-assignment limitation that VCBM's blend mechanism (Section~\ref{sec:technical-contribution}) directly targets.

This uniform advantage is consumed by a PPO-clipped policy-gradient objective, needed because rollouts are generated by a separate, possibly numerically divergent inference engine. The update requires an importance weight correcting for the mismatch between the sampling log-probability $\pi_{\theta_{\mathrm{old}}}$ and the current training log-probability $\pi_\theta$:
\begin{equation}
  w_t = \exp\big(\log \pi_\theta(a_t) - \log \pi_{\theta_{\mathrm{old}}}(a_t)\big) .
  \label{eq:importance-weight}
\end{equation}
The per-token objective applies a PPO-style clip \cite{b_schulman2017ppo} to bound any single update,
\begin{equation}
  \mathcal{L}_{\mathrm{PPO}}(t) = -\min\Big(w_t A_t,\; \mathrm{clip}(w_t,\, 1-\epsilon,\, 1+\epsilon)\, A_t\Big) ,
  \label{eq:ppo-clip}
\end{equation}
with clip range $\epsilon$ (default $0.1$ here). Equation~\eqref{eq:ppo-clip} is the loss into which VCBM's retrieval-weighted, per-token advantage (Section~\ref{sec:vcbm-appworld}) is substituted in place of the uniform $A_t = A^{(k)}$: VCBM modifies \emph{which} advantage each token receives, with no change to the surrounding PPO-clipped machinery.

\section{Related Work: Structural Alternatives to Uniform Credit}
\label{sec:related-work}

This section situates VCBM against the wider credit-assignment literature that motivates its design but is not directly benchmarked in Section~\ref{sec:experiments}, in four groups: temporal/counterfactual methods beyond RUDDER; group-relative/trajectory-level RL fine-tuning beyond LOOP; episodic memory and retrieval for LLM agents; and surveys of the problem's scope.

\subsection{Temporal and counterfactual credit assignment beyond RUDDER}
\label{sec:rw-temporal}

RUDDER (Section~\ref{sec:prelim-rudder}) belongs to a broader family that redistributes reward by learning an auxiliary predictive model rather than testing for causal structure directly. Hindsight Credit Assignment (HCA) \cite{b_hca2019} learns a hindsight distribution over which action, conditioned on an observed outcome, was responsible, and rewrites the value function accordingly, a \emph{learned, statistical} redistribution like RUDDER's LSTM. Counterfactual Credit Assignment (CCA) \cite{b_cca2021} conditions a value function on future events via a learned future-conditional baseline; COCOA \cite{b_cocoa2023} builds on this but measures contribution against a learned representation of the rewarding \emph{event} rather than the rewarding state, reducing HCA's spurious-credit failure mode. CCA and COCOA are conceptually closer to VCBM's causal framing, since both pose a counterfactual question, but both require a learned forward/counterfactual model to answer it, whereas VCBM's tabular test (Equation~\eqref{eq:causal-role}) needs only online frequency counts, trading generality for a test that is exact and directly verifiable (Section~\ref{sec:exp-watchrepair}). Randomized Return Decomposition \cite{b_randomized_return_decomp2022} and Counterfactual Shapley Credit Assignment \cite{b_phi_ppo2026} (a Monte-Carlo subsequence-sampling objective and a Shapley-value attribution grounded in a structural causal model, respectively) likewise redistribute credit continuously over every timestep; VCBM's departure is that a timestep either is or is not a bookmark, with no residual credit in between.

\subsection{Group-relative and trajectory-level RL fine-tuning beyond LOOP}
\label{sec:rw-trajectory}

LOOP's leave-one-out advantage (Section~\ref{sec:prelim-loop}) is one instance of a broader class computing a single scalar advantage per trajectory, applied uniformly to every token. Turn-level reward design for multi-turn reasoning \cite{b_multi_turn_grpo2025} extends GRPO/PPO with a dense, per-turn reward: the same coarse-granularity limitation as VCBM addresses, but by \emph{redesigning the reward} at every turn rather than retrieving and weighting a discrete subset afterward. G2PO \cite{b_g2po2026} is closer, being evaluated on AppWorld directly: it builds a global state-transition graph across rollouts and standardises temporal-difference errors on it to identify progress-driving transitions, exploiting state-transition structure much as VCBM's bookmark rule does. The distinction is mechanism: G2PO reads credit off a learned, cross-rollout graph, whereas VCBM's AppWorld instantiation uses a simple structural heuristic (state-mutating tool call) plus flat embedding-similarity retrieval, trading G2PO's richer structure for a mechanism cheap enough to run inline with no graph construction. SPA-RL \cite{b_spa_rl2025} densifies rewards via a learned stepwise-progress estimator, non-uniform like VCBM, but attributing \emph{progress} at every step rather than selecting a discrete causal subset; VCBM's bookmark mechanism is sparse where SPA-RL's is dense.

\subsection{Episodic memory and retrieval in long-horizon LLM agents}
\label{sec:rw-memory}

VCBM's retrieval mechanism (Equation~\eqref{eq:retrieval-softmax}) draws on using external memory to inform credit assignment. Memory-R2 \cite{b_memoryr22026} is the closest comparison: it addresses fair credit for memory-augmented LLM agents whose rollouts diverge because they write, update, or delete different memories, via LoGo-GRPO, which combines a global trajectory-level objective with local rerollouts from shared intermediate memory states, architecturally close to VCBM's blend of a bookmark memory with a trajectory-level advantage (Equation~\eqref{eq:vcbm-blend}). The distinction: LoGo-GRPO's fairness correction operates \emph{within} the group-relative advantage computation, changing which rollouts are compared, whereas VCBM keeps memory-weighting and the LOOP objective as two separable stages joined only by $\beta$. Memory-R2's setting is also explicitly \emph{multi-session}, while VCBM's AppWorld instantiation retrieves within/across single-episode rollouts of one training run, a related but not identical failure mode; this paper does not claim VCBM is a special case of LoGo-GRPO, only that both belong to the same family of memory-conditioned credit assignment.

\subsection{Surveys establishing the scope of the credit-assignment problem}
\label{sec:rw-surveys}

Four surveys help frame where VCBM sits in the literature. Pignatelli et al.\ \cite{b_pignatelli2023survey} survey temporal credit assignment in deep RL broadly, with a formalism this paper's tabular experiment is consistent with in spirit. Two further surveys narrow the scope to LLM agents: one \cite{b_agentic_rl_survey2026} identifies credit assignment as a central bottleneck for scaling LLM agents to long-horizon settings, and a companion \cite{b_agentic_ca_survey2026} organises 47 methods by granularity (token- to multi-agent-level) and methodology, noting that outcome-only signals grow uninformative as horizon lengthens. A fourth \cite{b_long_horizon_agents_survey2025} reviews the full long-horizon agent lifecycle, situating credit assignment as one stage among several. Together they indicate the field has largely converged on \emph{learned} solutions to the granularity problem that VCBM instead addresses, in its tabular instantiation, with an explicit statistical test.

\subsection{VCBM's distinguishing position}
\label{sec:rw-position}

Across all three groups above, most surveyed methods assign credit through a \emph{learned or continuously-valued} mechanism. VCBM's two instantiations are deliberately harder-edged: in the tabular case a bookmark either is or is not causally controlled (fixed-threshold hypothesis test, no smoothing); in the AppWorld case what counts as a bookmark is a structural rule rather than a learned classifier, though \emph{weighting} is soft (Equation~\eqref{eq:retrieval-softmax}), closest to memory-augmented methods like Memory-R2. CCA and COCOA share VCBM's causal framing without its hardness, substituting a learned counterfactual model. This paper's contribution is thus best read not as introducing causal or memory-based credit assignment as a category, but as testing how far a hard, discovery-or-heuristic-driven bookmark selection can go before the softer, learned mechanisms that dominate the literature become necessary.

\section{Visual Causal-chain Bookmarking}
\label{sec:technical-contribution}

Both VCBM instantiations share a three-stage template (\emph{detect} decision points that plausibly carry causal weight, \emph{store} them in structured memory, and \emph{weight} the learning signal accordingly), but the mechanism at each stage differs between a fully observed tabular MDP, where causal discovery is statistically tractable, and a partially observed, language-based MDP, where it is not. This section presents both in full and states explicitly where they diverge.

\subsection{VCBM for tabular MDPs: causal discovery and Welford estimation}
\label{sec:vcbm-tabular}

\subsubsection{Why causal discovery is tractable here}
In a fully observed MDP with a small, fixed-dimensional state space (the kind of structured representation that contrastive \cite{b_kipf2020cswm} and recurrent \cite{b_ha2018worldmodels} world models otherwise learn from raw observations), it is possible to test directly, from interaction data alone, which state components an action causally influences. Credit for the final return can then be assigned exactly to the decision points where a controlled change originates, rather than diffused across an entire trajectory of filler steps, as in the watch-repair environment of Section~\ref{sec:experiments}, where a single decision at $t=0$ determines the return realised 50 steps later.

\subsubsection{Causal-role discovery}
Let the state decompose into $n$ scalar components $s = (s^{(1)}, \dots, s^{(n)})$. For each component $i$ and action $a \in \{0,1\}$, a running count $c_{i,a}$ tracks how often $i$ changed value on a step with action $a$, out of $n_a$ such steps. After a warm-up period, each component is classified by its empirical change rate $\rho_{i,a} = c_{i,a}/n_a$:
\begin{equation}
  \mathrm{role}(i) =
  \begin{cases}
    \textsc{clock}      & \min(\rho_{i,0}, \rho_{i,1}) > \eta_{\mathrm{clock}} \\
    \textsc{static}     & c_{i,0} + c_{i,1} = 0                                \\
    \textsc{controlled} & z_i > z_{\mathrm{thresh}}                            \\
    \textsc{noise}      & \text{otherwise,}
  \end{cases}
  \label{eq:causal-role}
\end{equation}
where $z_i$ is the two-proportion test statistic comparing the change rate under each action,
\begin{equation}
  z_i = \frac{|\rho_{i,0} - \rho_{i,1}|}{\sqrt{\hat{p}(1-\hat{p})\big(1/n_0 + 1/n_1\big)}}, \quad \hat{p} = \frac{c_{i,0}+c_{i,1}}{n_0+n_1} .
  \label{eq:z-test}
\end{equation}
A component is \textsc{controlled} only if its change rate depends significantly on the action; \textsc{clock}/\textsc{static} components change deterministically regardless of action; \textsc{noise} components change at a rate independent of the action, and are excluded from the value-estimation key below. This test lets VCBM recover the watch-repair environment's causal structure (Section~\ref{sec:experiments}) without hand-specified labels. In our implementation $z_{\mathrm{thresh}}=4.0$ and $\eta_{\mathrm{clock}}=0.9$, and the classifier is not consulted until a warm-up period of 2{,}000 transitions has elapsed, after which its labels are refreshed every 500 further transitions.

\subsubsection{Bookmark discovery and value estimation}
A state is a \emph{bookmark opportunity} if its projection onto \textsc{controlled} and \textsc{clock} components matches a context previously observed immediately before a \textsc{controlled} component changed. At each bookmark, a key is formed from the \textsc{static}, \textsc{controlled}, and \textsc{clock} components only (marginalising out \textsc{noise}), and a Welford online estimator \cite{b_welford1962method} accumulates, per (key, action), the mean and variance of the \emph{full} episode return $G(\tau)$:
\begin{equation}
  \begin{aligned}
    \mu_n &= \mu_{n-1} + \frac{G(\tau) - \mu_{n-1}}{n}, \\
    M_{2,n} &= M_{2,n-1} + \big(G(\tau)-\mu_{n-1}\big)\big(G(\tau)-\mu_n\big) ,
  \end{aligned}
  \label{eq:welford}
\end{equation}
with standard error $\mathrm{SE} = \sqrt{M_{2,n}/(n(n-1))}$. Marginalising out noise means samples from episodes that differ only in noise trajectory are correctly pooled, sample sharing unavailable to a plain per-state value table.

\subsubsection{Action selection}
At a bookmark, actions are sampled uniformly until both reach a forced minimum of 60 visits each; between that floor and $n_{\min}=8$ subsequent visits, a two-sample $z$-test on the Welford means, $z = |\mu_0 - \mu_1| / \sqrt{\mathrm{SE}_0^2 + \mathrm{SE}_0^2}$, determines significance against a confidence threshold of $2.0$. If significant, the higher-mean action is chosen greedily (residual exploration $\epsilon_{\min}=0.005$); otherwise an $\epsilon$-greedy rule with $\epsilon$ decaying exponentially from $0.20$ at a per-episode rate of $0.995$ down to the same $\epsilon_{\min}$ floor continues exploring, in place of a goal-directed exploration curriculum \cite{b_florensa2018goalgan,b_mendonca2021lexa} unneeded in a two-action setting. At non-bookmark timesteps, an action is drawn uniformly, since Equation~\eqref{eq:causal-role} guarantees such steps do not causally affect the outcome once the classifier has converged.

\subsection{VCBM for language-based MDPs: retrieval-weighted credit blending}
\label{sec:vcbm-appworld}

\subsubsection{Why causal discovery breaks down at scale}
The causal-role test of Section~\ref{sec:vcbm-tabular} requires a fixed, low-dimensional, fully observed state whose components can be tracked individually. An LLM agent (a Transformer-based \cite{b_vaswani2017attention} policy) operating in AppWorld \cite{b_trivedi2024appworld} instead observes unstructured natural-language tool outputs from a partially observed environment, of the kind addressed more generally by deep variational RL for POMDPs \cite{b_igl2018dvrl}, with no fixed component decomposition and no tractable per-component hypothesis test online. The second instantiation therefore replaces statistical causal \emph{discovery} with a structural \emph{heuristic} for what counts as a bookmark, and exact-match lookup with similarity-based \emph{retrieval}, since no two AppWorld episodes share an identical state trajectory.

\subsubsection{Bookmark detection and episodic memory}
A turn is bookmarked if it issues a state-mutating tool call, detected via the environment server's execution interface with no extra network calls. Each bookmarked turn's tool-output text $x$ is embedded with a deterministic, dependency-free hashing-trick bag-of-words encoding $g_\phi$ into a fixed-dimension vector,
\begin{equation}
  g_\phi(x)_j = \frac{1}{\lVert v \rVert_2} \sum_{\mathrm{token} \in x} \mathbb{1}\big[h(\mathrm{token}) = j\big] ,
  \label{eq:hashing-encoder}
\end{equation}
where $h(\mathrm{token}) = \mathrm{int}(\mathrm{MD5}(\mathrm{token})[{:}8]) \bmod d$, and the tuple $(\mathrm{turn\_idx}, g_\phi(x), G(\tau))$ (position, embedding, and trajectory return) is stored in episodic memory. In our implementation $d=256$, tokens are extracted from the lowercased turn text with a simple alphanumeric regular expression, and $x$ is specifically the raw execution-result text returned by the interactive Python read-eval-print loop after a state-mutating call, i.e.\ the tool's own output rather than the agent's code or reasoning that produced it. If an episode contains no more than $k$ bookmarked turns, retrieval is skipped entirely and every bookmarked turn instead receives the flat floor weight $\alpha$, since retrieving the top-$k$ of fewer than $k$ candidates is degenerate.

\subsubsection{Retrieval and temperature-scaled weighting}
Given a trajectory's terminal-observation embedding $z_H = g_\phi(x_H)$, credit is distributed across the top-$k$ most similar bookmarks by cosine similarity: dense retrieval in the family of dense passage retrieval for text \cite{b_karpukhin2020dpr}, but over bookmarked trajectory turns rather than a text corpus,
\begin{equation}
  \mathrm{sim}(z_H, k_i) = \frac{z_H \cdot k_i}{\lVert z_H \rVert \, \lVert k_i \rVert}, \qquad
  w_i = \frac{\exp(\mathrm{sim}_i / \tau)}{\sum_{j \in \mathrm{top}\text{-}k} \exp(\mathrm{sim}_j / \tau)} ,
  \label{eq:retrieval-softmax}
\end{equation}
with temperature $\tau$ controlling how sharply weight concentrates on the most similar bookmark. Tokens in a retrieved bookmarked turn receive weight $w_i$; all other tokens receive a fixed floor weight $\alpha \in (0,1)$, down-weighted but never excluded; the weight vector is renormalised to sum to the output-token count, preserving overall gradient scale.

\subsubsection{Blending with the trajectory-level LOO advantage}
Rather than replacing LOOP's leave-one-out advantage outright, the per-token VCBM weight $\tilde{w}_t$ (Equation~\eqref{eq:retrieval-softmax}, renormalised) is combined with the uniform LOO advantage $A^{(k)}$ via a mixing coefficient $\beta \in [0,1]$:
\begin{equation}
  A_{\mathrm{VCBM},t} = A^{(k)} \cdot \tilde{w}_t, \qquad
  A_{\mathrm{blend},t} = (1-\beta)\, A^{(k)} + \beta\, A_{\mathrm{VCBM},t} ,
  \label{eq:vcbm-blend}
\end{equation}
which reduces exactly to plain LOOP at $\beta=0$ and to a fully bookmark-concentrated advantage at $\beta=1$; $A_{\mathrm{blend},t}$ substitutes directly for the uniform $A_t$ in Equation~\eqref{eq:ppo-clip}. Rollouts with zero bookmarked turns receive the unmodified scalar LOO advantage, since Equation~\eqref{eq:vcbm-blend} is undefined without a retrieval candidate.

\subsection{Where the two instantiations agree and where they necessarily diverge}
\label{sec:vcbm-comparison}

Both instantiations follow the detect/store/weight template and share the hypothesis that concentrating credit on causally relevant decision points outperforms an undifferentiated trajectory-level signal, but they are not the same algorithm. Table~\ref{tab:vcbm-comparison} makes the divergence explicit: the tabular instantiation \emph{discovers} bookmarks via a statistical test, with no learned embedding or similarity search; the AppWorld instantiation \emph{assumes} a structural rule and relies on embedding-space retrieval to generalise across episodes that never share an identical state. The two are two different answers to the same design question (how to concentrate credit when discovery is or is not statistically tractable), evaluated independently in each regime, rather than one method validated twice.

\begin{table}[htbp]
  \centering
  \caption{Mechanistic comparison of the two VCBM instantiations.}
  \label{tab:vcbm-comparison}
  \begin{tabular}{@{}p{1.7cm}p{2.9cm}p{2.9cm}@{}}
    \toprule
    \textbf{Aspect} & \textbf{Tabular} & \textbf{AppWorld/LLM} \\
    \midrule
    Bookmark detection & Statistical causal-role test (Eq.~\eqref{eq:causal-role}) & Structural rule (state-mutating tool call) \\
    Memory content & Welford $(\mu, \sigma^2)$ per (key, action) & Raw (index, embedding, return) tuples \\
    Retrieval mechanism & Exact key match & Cosine-similarity top-$k$ (Eq.~\eqref{eq:retrieval-softmax}) \\
    Underlying policy & Tabular $Q$-style controller & LoRA-adapted LLM, PPO-clipped (Eq.~\eqref{eq:ppo-clip}) \\
    Fallback when scarce & Balanced random sampling & Flat floor weight $\alpha$ \\
    \bottomrule
  \end{tabular}
\end{table}

\subsection{Six structural capabilities VCBM has that a uniform trajectory-level baseline lacks}
\label{sec:vcbm-capabilities}

The detect/store/weight template gives both instantiations six capabilities LOOP's leave-one-out advantage (Equation~\eqref{eq:loo-advantage}) structurally lacks, being a single scalar $A^{(k)}$ applied uniformly with no notion of \emph{which} token mattered:
\begin{itemize}
  \item \emph{Fine-grained credit}: the causal-role test (Eq.~\eqref{eq:causal-role}) and AppWorld bookmark rule identify specific credit-bearing timesteps, expected to lower gradient variance and reduce reinforcement of spurious correlations.
  \item \emph{Per-decision value estimates}: Welford (Eq.~\eqref{eq:welford}) and retrieval-softmax (Eq.~\eqref{eq:retrieval-softmax}) key weight to an individual (bookmark, action) or turn, not the whole rollout, useful where one decision is pivotal, as in watch-repair's $t{=}0$ choice.
  \item \emph{Causal/structural knowledge}: only the tabular case has this genuinely (Eq.~\eqref{eq:causal-role} separates \textsc{controlled} from independently-changing components); AppWorld substitutes a weaker, assumed heuristic. Uniquely verifiable against ground truth (Section~\ref{sec:exp-watchrepair}).
  \item \emph{Persistent memory}: the Welford accumulator and AppWorld memory bank carry information across iterations, unlike LOOP's baseline, recomputed and discarded each batch.
  \item \emph{Retrieval-based generalisation} (AppWorld only): Eq.~\eqref{eq:retrieval-softmax} matches a trajectory to similar past bookmarks by cosine similarity rather than only the $K-1$ others in its batch, most useful where exact scenario repeats are rare.
  \item \emph{Selective, weighted credit}: the tabular key marginalises out \textsc{noise}; the AppWorld blend concentrates weight on bookmarks while down-weighting (not eliminating) the rest via floor $\alpha$, deliberately non-uniform, unlike LOOP's identical $A^{(k)}$ per token.
\end{itemize}
These are design rationale, not uniformly confirmed results. Only the first three are empirically observed, and only in the tabular setting: VCBM's lower-mean, lower-variance episodes-to-converge (Table~\ref{tab:watchrepair-results}) and exact causal-structure recovery are confirmed significant (Mann--Whitney $U=19$, $p<0.001$, $r=0.91$). The remaining three are structurally present in AppWorld by construction and directionally consistent with the single-seed comparison (Section~\ref{sec:exp-appworld}), but not yet statistically confirmed there; a second-seed AppWorld run (Section~\ref{sec:conclusions}) would be the natural next step to test them.

\section{Experiments: Tabular Ground Truth and AppWorld Scale-Up}
\label{sec:experiments}

\subsection{Controlled evaluation: RUDDER vs.\ VCBM in a delayed-reward tabular MDP}
\label{sec:exp-watchrepair}

\subsubsection{Environment}
The watch-repair environment models a single consequential decision under delayed, stochastic feedback. The state is a 5-tuple (repaired-flag, two transport-condition counters, brand, timestep); at $t=0$ the agent observes a random watch \textsc{brand} $b \sim \mathrm{Uniform}\{0,1,2,3\}$ (uniform appearance probability across all four brands in the reference implementation) and chooses to repair it ($a=0$, stochastic cost $r^{\mathrm{repair}} \sim \mathcal{N}(\mu_{\mathrm{repair}}[b], \sigma^2_{\mathrm{repair}}[b])$, with per-brand means $\mu_{\mathrm{repair}} = [1, 4, 5, 24.5]$ and variances $[2, 2, 2, 1]$) or not ($a=1$, no cost). For the remaining $T=50$ timesteps, two independent \textsc{noise} counters increment as Bernoulli trials with fixed, action-independent per-step probabilities $[0.10, 0.05]$, and a deterministic \textsc{clock} advances; the action has no further effect after $t=0$. Only a repaired watch yields a terminal reward at $t=50$: the brand's resale price ($[18, 28, 31, 59]$ for brands 0--3) minus a stochastic brand-related transport cost ($\mathcal{N}(\mu, \sigma^2)$ with per-brand means $[0.5, 2.5, 4.5, 18]$, shared variance $1.5$) and the accumulated transport-condition penalty (each counter's final value multiplied by a per-unit cost of $0.1$ and $7$ respectively). Because repair cost, resale price, and transport-cost sensitivity all vary by brand, the return-maximising policy is brand-dependent: repairing is optimal only for brands 1 and 2, a known ground truth used purely for evaluation, never training.

\subsubsection{Method and metrics}
Both RUDDER (Section~\ref{sec:prelim-rudder}, redistributing reward into a tabular direct-$Q$ update) and VCBM (Section~\ref{sec:vcbm-tabular}) were trained for 20 independent seeds, terminated once the moving average (750-episode window) fraction of optimal first-decisions reached 95\%. Table~\ref{tab:watchrepair-results} reports episodes-to-convergence, our primary sample-efficiency metric \cite{b_levine2020offlinerl}, alongside converged decision quality.

\begin{table}[htbp]
  \centering
  \caption{Watch-repair environment: episodes to reach 95\% correct first-decisions, 20 seeds each.}
  \label{tab:watchrepair-results}
  \begin{tabular}{@{}lp{1.6cm}p{1.9cm}c@{}}
    \toprule
    \textbf{Method} & \textbf{Range} & \textbf{Mean $\pm$ std} & \textbf{Final \%} \\
    \midrule
    RUDDER & 1{,}603--10{,}936 & 3{,}327.1 $\pm$ 2{,}010.9 & 95.07\% \\
    VCBM   & 1{,}196--7{,}757  & 1{,}532.3 $\pm$ 1{,}428.0 & 95.07 $\pm$ 0.00\% \\
    \bottomrule
  \end{tabular}
\end{table}

\subsubsection{Discovered causal structure}
Across all 20 VCBM seeds, the causal-role classifier (Equation~\eqref{eq:causal-role}) recovered an identical label assignment for the five state components (repaired-flag, two transport counters, brand, timestep): \textsc{controlled}, \textsc{noise}, \textsc{noise}, \textsc{static}, \textsc{clock}. This exactly matches the environment's true generative structure: only the repaired-flag's change rate depends on the action, the transport counters change at fixed rates, the brand is fixed, and the timestep is deterministic. This is direct evidence that VCBM's discovery mechanism recovers ground-truth causal structure from interaction data alone, without being told which of the five components matters.

\subsubsection{Interpretation: a confirmed effect}
Both methods reach 95\% in all 20 seeds, but VCBM's episodes-to-converge distribution has a lower minimum (1{,}196 vs.\ 1{,}603) and maximum (7{,}757 vs.\ 10{,}936) than RUDDER's, consistent with concentrating the signal on the causally relevant decision point converging at least as fast, with less tail risk, than an LSTM-based redistribution over the full trajectory. Because both distributions are right-skewed and non-normal, an unpaired two-sided Mann--Whitney $U$ test (normal approximation, tie correction) rejects the null that the two distributions are identical ($U=19$, $z=-4.90$, $p<0.001$), with a large effect size ($r=0.91$), confirming VCBM's lower episodes-to-converge is statistically significant rather than sample noise.

\begin{figure}[htbp]
  \centering
  \includegraphics[width=0.85\linewidth]{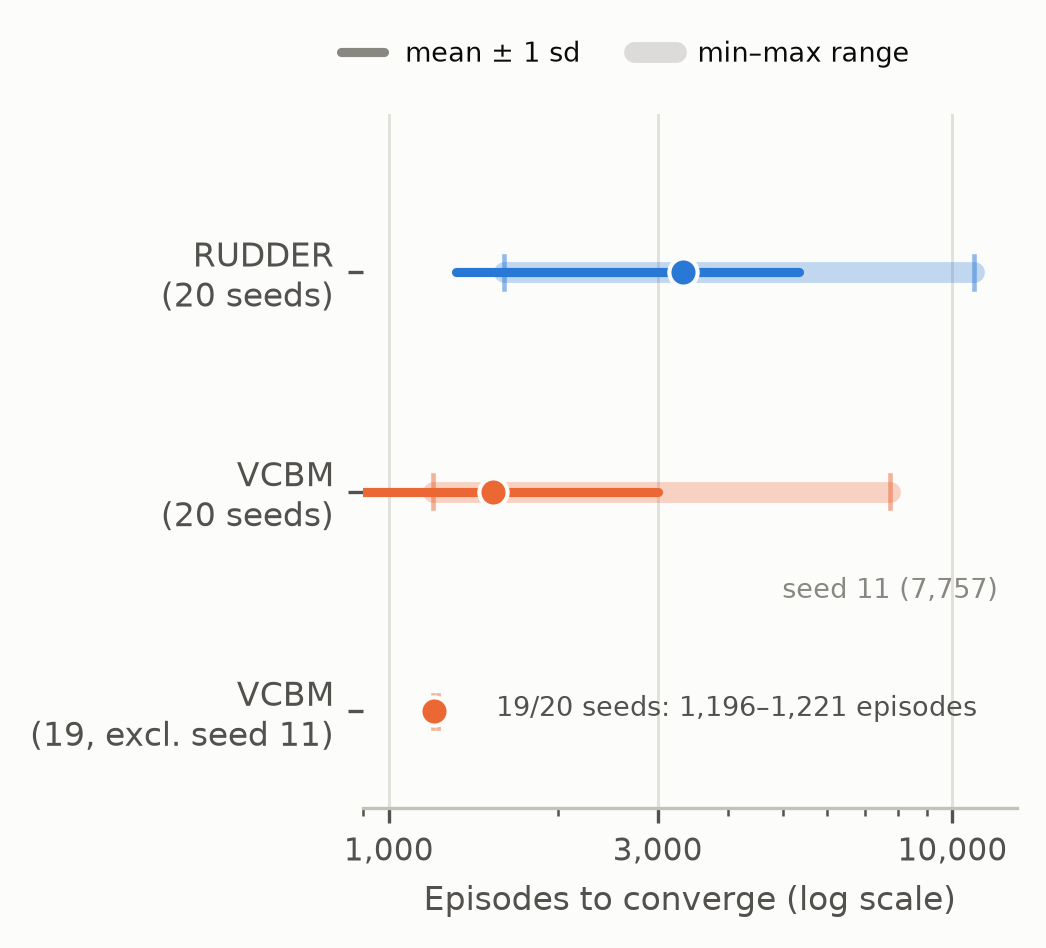}
  \caption{Episodes to converge, RUDDER vs.\ VCBM, 20 seeds each (log-scale axis). Each row shows the full min--max range (light band) and the mean $\pm$ 1 standard deviation (dark segment, marker at the mean). Excluding a single outlier seed (seed 11, 7{,}757 episodes), the remaining 19 VCBM seeds converge within a 25-episode band (1{,}196--1{,}221), against RUDDER's far wider spread across all 20 seeds.}
  \label{fig:watchrepair-reproducibility}
\end{figure}

\subsection{Large-scale evaluation: LOOP vs.\ VCBM-blended LOOP on AppWorld}
\label{sec:exp-appworld}

\subsubsection{Training setup and the credit-mechanism ablation}
Both runs share Qwen2.5-7B-Instruct as the base policy, served via vLLM with prefix caching and a 32{,}768-token context, LoRA \cite{b_hu2022lora} rank 8 / alpha 16 (dropout 0) on all seven attention and MLP projections, bf16 precision, FSDP2 training across 2 GPUs (a single-node instance of the distributed actor-learner pattern used at larger scale by IMPALA \cite{b_espeholt2018impala}), AdamW \cite{b_kingma2015adam} at a constant learning rate of $5\times10^{-5}$ (fused, weight decay $0.01$, no warm-up) with gradient-norm clipping at $0.1$, 200 iterations, the AppWorld train split, sparse per-episode reward, and up to 50 interactions per episode capped at 1{,}500 generated tokens each. The agent follows a ReAct-style prompting loop, one fixed one-shot worked example followed by task instructions, with prompts truncated in three stages (blanking old observation blocks, then truncating individual long observations, then dropping the oldest blocks) once the running prompt exceeds 20{,}000 characters. Training discards a per-token advantage update if the estimated KL divergence for that iteration exceeds a high-KL abort threshold of 20 nats. The only intended difference is the credit mechanism: the baseline (\texttt{csf3\_7b\_2gpu\_90iter}) uses plain LOOP ($\beta=0$), and the comparison (\texttt{csf3\_7b\_2gpu\_90iter\_vcbm\_r2}) uses VCBM-blended LOOP ($\beta=1$, $\tau=0.5$, top-$k=6$, $\alpha=0.05$).

\emph{Provenance and confidence caveat.} Final-logged-step figures below are exact, from the Weights \& Biases (W\&B) summary export for each run. Run-peak figures (Table~\ref{tab:appworld-drift}) and outlier-gradient timing are chart-read from each run's full per-iteration W\&B dashboard export (223$\times$213-pixel rendered tiles, no numeric time-series export available), so they carry an estimated $\pm$10--20\% reading uncertainty and are marked $\sim$. Exact claims are single-seed, final-step results; chart-read claims are directionally informative but not precise; neither is statistically tested in the sense of Section~\ref{sec:exp-watchrepair}'s Mann--Whitney test.

\emph{Why single-seed.} Each 200-iteration, 9{,}600-rollout run occupies 2 GPUs for its full duration; the tabular experiments (Section~\ref{sec:exp-watchrepair}) are cheap enough to run 20 seeds each on commodity hardware, but a single AppWorld run was already at the edge of the compute available for this project, and a second seed per arm was not feasible within that budget. This is a resource constraint, not a design choice: the tabular result is where the causal claim is established with a proper significance test across seeds; the AppWorld result is reported as a single-run demonstration that the same mechanism transfers to a harder, partially observed, natural-language setting, and is scoped accordingly throughout this section.

\subsubsection{Task performance}
Table~\ref{tab:appworld-results} reports the headline task-completion metrics: mean episode return and success rate, both overall and split by AppWorld's three difficulty tiers.

\begin{table}[htbp]
  \centering
  \caption{AppWorld task performance, final logged training step.}
  \label{tab:appworld-results}
  \begin{tabular}{@{}p{2.7cm}ccc@{}}
    \toprule
    \textbf{Metric} & \textbf{LOOP} & \textbf{VCBM} & \textbf{Change} \\
    \midrule
    Mean return             & 0.6682 & 0.7300 & +9.3\%    \\
    Success, all diff.\     & 0.3542 & 0.4583 & +29.4\%   \\
    Success, diff.-avg.\    & 0.2569 & 0.3333 & +29.7\%   \\
    Success, difficulty 1   & 0.5833 & 0.5000 & $-$14.3\% \\
    Success, difficulty 2   & 0.1875 & 0.5000 & +166.7\%  \\
    Success, difficulty 3   & 0.0000 & 0.0000 & no change \\
    \bottomrule
  \end{tabular}
\end{table}

Both runs solve zero difficulty-3 tasks ($\approx$17\% of rollouts). The gain is concentrated almost entirely in difficulty 2 (+166.7\%); difficulty 1 is unexpectedly slightly \emph{lower} for VCBM ($-$14.3\%), the only metric favouring LOOP.

\begin{figure}[htbp]
  \centering
  \includegraphics[width=\linewidth]{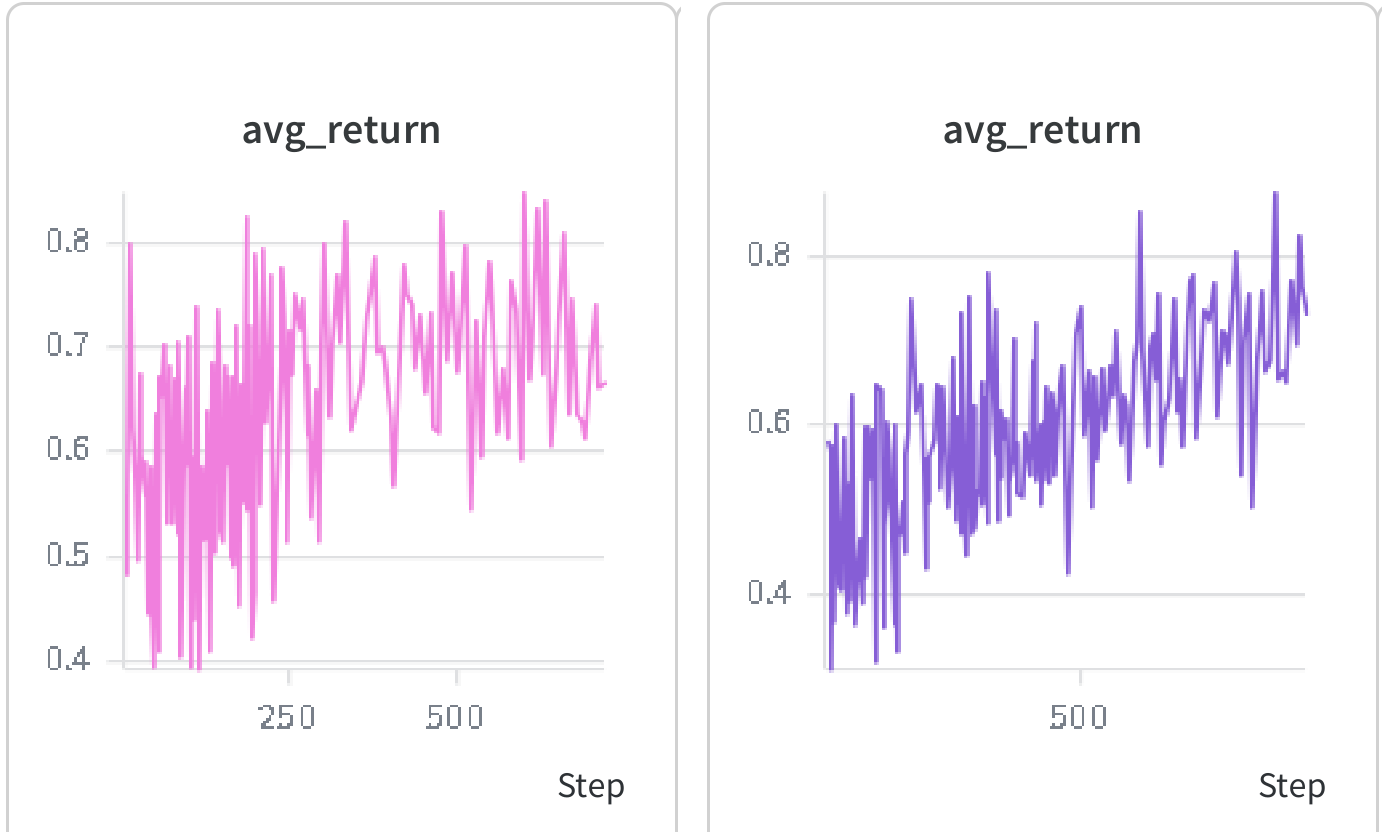}
  \caption{Mean rollout return per training iteration, single seed each. Left: plain LOOP. Right: VCBM-blended LOOP. Both trend upward over training; VCBM's later-training returns cluster higher and its trace extends further in logged iterations. Raw W\&B panels, not independently smoothed.}
  \label{fig:appworld-return}
\end{figure}

\subsubsection{Off-policy drift and optimisation stability}
Because \texttt{epochs\_per\_iteration=2} with \texttt{strictly\_on\_policy=false}, each rollout batch is used for two gradient passes, so the second is off-policy w.r.t.\ the rollout-generating weights; the importance weight $w_t$ (Section~\ref{sec:prelim-loop}) quantifies the resulting policy drift. Table~\ref{tab:appworld-drift} reports the maximum importance weight and KL estimate at the final logged step (exact) alongside their run-peak value (chart-read; see the provenance caveat above).

\begin{table}[htbp]
  \centering
  \caption{AppWorld off-policy drift and stability indicators.}
  \label{tab:appworld-drift}
  \begin{tabular}{@{}p{2.7cm}ccc@{}}
    \toprule
    \textbf{Metric} & \textbf{LOOP} & \textbf{VCBM} & \textbf{Change} \\
    \midrule
    Max imp.\ wt., final    & 3.97       & 3.09       & $-$22.2\%     \\
    Max imp.\ wt., peak     & $\sim$49   & $\sim$14   & $\sim-$71\%   \\
    Max log-prob.\ diff., peak & $\sim$12.5 & $\sim$5.5  & $\sim-$56\% \\
    KL estimate, final      & 0.561      & 0.583      & +4.1\%        \\
    KL estimate, peak       & $\sim$18.5 & $\sim$18.5 & $\sim$ tied   \\
    Outlier-grad.\ events   & 2          & 2          & tied          \\
    \bottomrule
  \end{tabular}
\end{table}

Mean probability drift is essentially identical between runs; the maximum importance weight is lower for VCBM at both the final step ($-$22.2\%) and run peak ($\sim$14 vs.\ $\sim$49), consistent with concentrating gradient mass on fewer tokens reducing worst-case drift. The KL estimate, however, is marginally higher for VCBM at the final step (+4.1\%) and essentially tied at the run peak (both $\sim$18.5, against the 20-nats abort threshold): a mixed rather than uniformly favourable result. Both runs recorded exactly two cumulative outlier-gradient events \cite{b_du2017saddlepoints,b_hardt2016trainfaster}; LOOP's second occurs around iteration $\sim$95--100, VCBM's later, $\sim$150--160 (chart-read).

\subsubsection{Efficiency}
Table~\ref{tab:appworld-efficiency} reports output-token usage and a derived reward-per-token efficiency figure, since shorter trajectories are only a genuine efficiency gain if success rate does not fall.

\begin{table}[htbp]
  \centering
  \caption{AppWorld generation efficiency, final logged training step.}
  \label{tab:appworld-efficiency}
  \begin{tabular}{@{}p{3.3cm}ccc@{}}
    \toprule
    \textbf{Metric} & \textbf{LOOP} & \textbf{VCBM} & \textbf{Change} \\
    \midrule
    Output tokens / rollout & 2{,}001.6 & 1{,}625.2 & $-$18.8\% \\
    Return / 1{,}000 tokens & 0.334     & 0.449     & +34.6\%   \\
    \bottomrule
  \end{tabular}
\end{table}

VCBM's token count is 18.8\% lower than LOOP's while its success rate is simultaneously higher (Table~\ref{tab:appworld-results}), so the reduction reflects solving tasks in fewer steps rather than terminating early. \emph{Return per 1{,}000 output tokens}, computed as \texttt{avg\_return}/(\texttt{avg\_output\_tokens}/1000), is 34.6\% higher for VCBM, the largest relative improvement in this section.

\subsubsection{Parity and open questions}
Both runs completed 200 iterations, 9,600 rollouts, the same base model, and the same learning rate. Diffing the exported W\&B configurations (213 fields) shows they are \emph{not} a clean single-field ablation: three VCBM-specific hyperparameters differ simultaneously ($\tau$: 1 vs.\ 0.5; top-$k$: 4 vs.\ 6; $\alpha$: 0.1 vs.\ 0.05), alongside $\beta=1$ for VCBM. Since LOOP does not consume these fields, this is not a confound, but a comparison against one specific choice of hyperparameters, not a sweep. Two discrepancies remain unresolved in LOOP's own configuration: its logged rollout count (48) does not follow arithmetically from its stated settings, and \texttt{llm.temperature=0} would imply zero advantage everywhere, contradicting the observed non-zero \texttt{adv\_filtered\_fraction}.

\subsubsection{Interpretation: a directional signal}
VCBM shows a higher mean return, higher success rate, lower maximum importance weight, and better token efficiency than LOOP; only the KL estimate does not favour it. As stated above, this is a single-seed, partly chart-read comparison, so none of these differences have been tested for significance the way the tabular result was, and none should be read as a confirmed effect. The two experiments in this paper are deliberately asymmetric in evidentiary weight: the tabular result (Section~\ref{sec:exp-watchrepair}) is a controlled, statistically confirmed test of the central causal-credit-assignment claim under full observability; the AppWorld result is a compute-limited demonstration that the same blend mechanism is at least compatible with an improvement in a harder, partially observed, natural-language setting, not an independent confirmation of it. A reader should weight the two accordingly.

\section{Conclusions: When Structural Credit Assignment Helps}
\label{sec:conclusions}

This paper tested whether RL credit assignment can be improved by explicitly identifying causally relevant decision points, rather than relying solely on trajectory-level baselines (LOOP) or learned return redistribution (RUDDER), via two instantiations jointly termed VCBM. In the tabular watch-repair environment, VCBM recovered the true causal structure across all 20 seeds and converged with a narrower, lower episode-count distribution than RUDDER, confirmed statistically significant (Mann--Whitney $U=19$, $p<0.001$, $r=0.91$; Section~\ref{sec:exp-watchrepair}), directly supporting the central hypothesis where ground truth is fully verifiable. In AppWorld, VCBM-blended LOOP showed a higher mean return, higher success rate, lower final-step maximum importance weight, and better token efficiency, against a marginally higher final-step KL estimate (Section~\ref{sec:exp-appworld}); these figures are exact but single-seed, so no significance test has been or can yet be run, and the AppWorld result is reported at the confidence the evidence supports: directionally consistent with the central hypothesis, but not an independent confirmation of it.

Three limitations remain, in order of how much they should temper the reader's confidence. First, and most substantively: the AppWorld comparison is a single seed per arm, with no significance test, because each 200-iteration/9{,}600-rollout run occupies 2 GPUs for its full duration and a second seed per arm was outside the compute budget available for this project. This is a genuine, compute-bound gap, not a claim to be argued around; it is why Section~\ref{sec:exp-appworld} is explicit that the AppWorld result carries less evidentiary weight than the tabular one and should be read as a scaling demonstration, not a controlled trial. Closing it needs either a second AppWorld seed under the same budget constraints (the natural next step, at roughly the same 2-GPU/run cost already spent once per arm here) or a cheaper proxy, such as a shorter-horizon AppWorld run or a reduced task subset, used only to bound the variance rather than to reproduce the full 200-iteration comparison. Second, the AppWorld bookmark detector is a hand-specified structural rule, not a learned criterion, so it may not generalise where causally relevant actions are not tool calls; a learned detector trained on the tabular classifier's data (Equation~\eqref{eq:causal-role}) would close this gap (Table~\ref{tab:vcbm-comparison}). Third, Memory-R2's LoGo-GRPO \cite{b_memoryr22026} also uses memory to shape credit for memory-augmented LLM agents, and while Section~\ref{sec:rw-memory} argues a mechanistic distinction, this has not been verified empirically. All three are left for future work.

\end{document}